\documentclass[preprint,12pt]{elsarticle}

\usepackage{amssymb}

\usepackage{xcolor}
\usepackage{booktabs}
\usepackage{amsmath,amssymb,amsthm}
\usepackage{amsfonts}
\newcommand{\bm}[1]{\mathbf{#1}}
\usepackage{multirow}
\usepackage{algorithmic}
\usepackage{algorithm}
\usepackage{float}
\usepackage[lofdepth,lotdepth]{subfig}
\usepackage{booktabs}
\usepackage{graphicx} 
\usepackage{wrapfig}
\usepackage{hyperref}
\journal{Neural Networks}
\usepackage{setspace}
\newcommand{\eat}[1]{}

\begin{document}

\begin{frontmatter}



\title{Unsupervised Domain-adaptive Hash for   Networks}

	 
\author[1]{Tao He}
\ead{tao.he@monash.edu}
\author[2]{Lianli Gao}
\ead{Lianli.gao@uestc.edu.cn}
\author[2]{Jingkuan Song}
\ead{Jingkuan.song@gmail.edu}
\author[1]{Yuan-Fang Li$^*$}
\ead{Yuanfang.li@monash.edu}
\ead{Corresponding Author.}

\address[1]{Department of Data Science and AI, Faculty of Information Technology, Monash University, Clayton, Victoria 3800 }
\address[2]{Center for Future Media, University of Electronic
	Science and Technology of China, Chengdu, Sichuan 611731  }

\begin{abstract}
{Abundant real-world data can be naturally represented by large-scale networks, which demands efficient and effective learning algorithms. 
At the same time, labels may only be available for some networks, which demands these algorithms to be able to adapt to unlabeled networks.} 
Domain-adaptive hash learning has enjoyed considerable success in the computer vision community in many practical tasks due to its lower cost in both retrieval time and storage footprint. However, it has not been applied to multiple-domain networks. 
In this work, we bridge this gap by developing an unsupervised domain-adaptive hash learning method for networks, dubbed UDAH.
Specifically, we develop four {task-specific yet correlated} components: (1) network structure preservation via a hard groupwise contrastive loss, (2) relaxation-free supervised hashing, (3) cross-domain intersected discriminators, and (4) semantic center alignment. We conduct a wide range of experiments to evaluate the effectiveness and efficiency of our method on a range of tasks including link prediction, node classification, and neighbor recommendation. Our evaluation results demonstrate that our model achieves better performance than the state-of-the-art conventional discrete embedding methods over all the tasks. 
\end{abstract}

 


\end{frontmatter}


	\section{Introduction}

\emph{Hash learning}, or learning to hash~\cite{wang2018survey}, aims at learning low-dimensional, discriminative and compact binary codes from high-dimensional features or attributes.
With the explosive increase in the amount of available digital data, hash learning has become a prominent approach to effective data compression, as it enables efficient storage (in terms of space) and fast retrieval (in terms of time). 
{Networks can naturally represent data in diverse real-world applications. As a result, representation learning (i.e., embedding) methods~\cite{zhang2018network}, especially those based on neural networks, have become an active research problem in the machine learning and deep learning communities.} 

For networks, hashing is useful for converting high-dimensional representations of nodes into semantical binary codes, taking into account neighborhood proximity. 
Although hashing techniques~\cite{shen2018discrete,wu2018efficient,yang2018binarized} have been explored for network embedding, most of them focus on single-domain networks, that is, these hash functions only work well on the source domain but would perform badly on a target (new) dataset with a large distributional shift from the source. 
{In other words, these methods do not adequately address the \emph{domain adaptation} problem in network hashing.}

A natural solution is to \emph{finetune}~\cite{tajbakhsh2016convolutional} the hash function learned on a source domain so that it can handle a target domain. However, the problem of finetuning lies in the fact that retraining the model on  new datasets requires availability of human annotations on the target domain, which is expensive to obtain and thus may not be available. 
 Therefore, it is critical to learn a domain-adaptive hash function that is able to handle multiple domains  without supervised training on the unlabeled target domain.
Recently, unsupervised domain adaptive learning~\cite{inproceedings2020,song2020unsupervised} has attracted significant attention. {The aim of this task is to transfer knowledge learned in the supervised setting on the source domain to the target domain, which is unlabeled.} 
However, the distribution discrepancy between the source and target domains becomes a main obstacle to the knowledge transfer. Specifically, that disparity could result in undesirable non-alignment of the two domains' embeddings on the common space, which heavily influences the prediction on the target domain. Hence, how to effectively align the two domains is a central challenge in domain adaptive learning. 

To alleviate the disparity issue, a suite of techiques~\cite{9241052,DBLP:conf/iccv/TzengHDS15,wu2020unsupervised,DBLP:conf/cvpr/TzengHSD17} employ Generative Adversarial Networks (GANs), and train a discriminator network aiming at judging whether the features comes from the source domain or the target domain. At the same time, a feature learning component tries to fool the discriminator so that it cannot distinguish the origin of the features.  When the discriminator cannot determine the origin of the features, domain disparity has been suppressed to a relatively low level. However, the discriminator in  \cite{9241052,wu2020unsupervised} is only aware of which domain the feature comes from, but unaware of the specific semantics of the feature, ultimately leading to  coarse alignment of tow domains,  \  on their common space. {Additionally, though the discriminators can distinguish the distribution of continuous features, they are not designed to judge the distribution of discrete codes~\cite{cao2018hashgan}}, which  limits the application of GANs in domain adaptive hash learning. 

Another major challenging facing domain adaptive hashing is the non-differentiable nature of hashing function~\cite{cao2017hashnet,do2016learning}, which is caused by the widely-used, non-differentiable \emph{sign} function. This problem has been mitigated by replacing it with the differentiable and continuous \emph{tanh} function. However, the tanh function could produce undesirable relaxation errors~\cite{DBLP:conf/iccv/TzengHDS15} and degrade the quality of learned hash codes. Although some works \cite{shen2018discrete,yang2020discrete} have proposed to leverage an alternating algorithm to optimize hash codes, it is hard to integrate it into a deep neural network in an end-to-end fashion.

In this work, we propose UDAH, an \textbf{U}nsupervised \textbf{D}omain-\textbf{A}daptive \textbf{H}ashing method for networks that effectively transfers knowledge learned on the source domain to the target domain. UDAH address the three issues discussed above: {(1) how to enable the  learned knowledge on source domain to transfer to the target domain;} (2) how to effectively  align two domains on the commone space; and (3) how to alleviate the issue of vanishing gradient produced by the relaxed hash function tanh. 
To address the first issue, we devise cross-domain intersected discriminators with a knowledge distillation loss. For the second, we explore a   semantic centers alignment component   
to  ensure that  semantic space   
are explicitly aligned.  {
Last, we adopt a reparameterization trick, i.e., Gumbel-Softmax~\cite{jang2016categorical}, which has enjoyed great success in other discretization tasks such as product quantisation~\cite{wu2019learning}. Gumbel-Softmax can effectively reduce gradient vanishing and enable our model to be trained in an end-to-end manner. }

In summary, our main contributions are fourfold.

\begin{enumerate}
	\item  We propose an unsupervised domain adaptive hashing method for networks, dubbed UDAH, which can  be trained on the source domain in a supervised fashion, and transfered to the unlabeled target domain. To the best of our knowledge, {we are the first to propose a technique that is dedicated to learning domain-adaptive hash functions for networks}. 

	\item {We develop two components, cross-domain discriminators 
	 and semantic centers alignment, to reduce domain distribution disparity. }
	
	\item To allow the model to effectively preserve neighborhood structure, we devise a hash groupwise contrastive loss, which shows superiority to the conventional pairwise constraint.

	\item We evaluate UDAH on three domain-adaptive networks. Our results strongly demonstrate that UDAH outperforms the other state-of-the-art  methods on three tasks: link prediction, node classification, and node recommendation. Furthermore, {we theoretically analyze  the feasibility of reducing domain disparity for UDAH }.
	
\end{enumerate}

{
A short conference version of this paper has appeared in IJCAI 2019~\cite{he2019one}. The preliminary version proposed a domain-adaptive hashing method for images, which leverages generative adversarial networks (GANs) to alleviate the distribution discrepancy. In this manuscript, we have made the following major extensions:}
\begin{enumerate}
   \item We first propose the task, domain-adaptive hash for networks, aiming to learn a transferable hash function so that it performs well on multiple-domain networks with large distribution disparity. 
   
   \item We further adopt a knowledge distillation strategy to reduce domains discrepancy of networks.   

   \item We leverage a reparameterization technique to design a differentiable hash function. 
   
   \item We propose a hard groupwise contrastive loss for network embedding. 
  
   \item We provide a theoretical analysis about the relationship between reducing domain discrepancy and UDAH. 
\end{enumerate}

\section{Related Work}

We briefly survey relevant literature from three aspects: domain-adaptive learning, network embedding, and learning to hash. 

\subsection{Domain Adaptation} 
The main purpose of domain adaptation is to manipulate  supervised information on the source domain to guide model training on the target domain without ground-truth labels. 
In the computer vision community, domain adaptation has been applied in segmentation~\cite{zhao2019multi} and image retrieval~\cite{Long:2018:DDA:3209978.3209999}. Unlike traditional training datasets that consist of single-domain data, 
domain-adaptive learning aims to train a unified model so that it can handle multiple domains (e.g., digits and handwritten numbers). The main challenge of domain adaptation lies in the distributional discrepancy between different domains, also named as domain shift in some other fields. 
To this end, many unsupervised strategies has been proposed to diminish the domain distribution semantical mismatch.
A fundamental idea is to supervisedly train a classifier on the source domain and then finetune it on the new domain~\cite{DBLP:journals/pami/BruzzoneM10}. Some works focus on how to assign high-confidence  labels, also named as pseudo labels, for the target domain. In this work~\cite{Hu_2018_CVPR}, an autoencoder based duplex networks was proposed to bridge the distribution gap between the two domains by equipping two generator networks with the ability to reconstruct both domains' data from their latent space. 
In fact, a standard practice of dealing with adaptive learning is to project the source and target domains into a common space and then reduce the domain disparity~\cite{DBLP:conf/iccv/FernandoHST13} by a distribution alignment component or loss function. 

\subsection{Network Embedding} 
Network embedding~\cite{zhang2018network} aims to map each node or edge in a network (e.g.,  heterogeneous network, homogeneous network, attributed network, etc) into a low-dimension vector and simultaneously preserve the network's information, including structure and semantics as much as possible. 

{Matrix Factorization} is widely adopted in many previous studies~\cite{goyal2018graph}. The core idea of matrix factorization is to construct a high-dimensional graph matrix, namely a Laplacian matrix or a similarity matrix, from {high dimensional data features} 
 and then use some dimensionality reduction strategies (e.g., SVD and NMF, etc.) to transform the original high dimension vectors to low-dimensional, compact structure-preserving embeddings. Specifically, \cite{cai2007spectral} first proposed an objective function based on Graph Laplacian Eigenmaps aiming to enable embedded vectors to approximate their original similarity matrix constructed by Laplacian transform. Due to the fact that the similarity matrix plays an important role in the embedding process, many subsequent works investigated how to construct an information-rich and representative matrix so that the embedded vectors  are equipped with more similarity information. 

\eat{Although DeepWalk~\cite{perozzi2014deepwalk}, LINE~\cite{tang2015line} does not explicitly factorize an affinity matrix, their essential optimization is equal to factorize a referring matrix~\cite{qiu2018network}. Specifically,
DeepWalk first explored a sequence model used in natural language processing to embedding a large graph and gained a significant performance in several tasks. Similarly, LINE proposed a general model to preserve the first-order and second-order proximity by a joint learning strategy and can handle different types of networks, especially large-scale networks. 
}

{Deep Learning} is a mainstream technique to conduct network embedding in recent years, especially with the boom of graph convolutional networks (GCN)~\cite{kipf2016semi}.
Autoencoders~\cite{pan2018adversarially} are a widely used technique in network embedding methods~\cite{he2021semisupervised} that are based on deep learning. The core goal of autoencoders is to bridge the gap between the input and output by an encoder and a decoder, where the  encoder aims to project the input data into a latent space by nonlinear mapping functions, and the decoder inversely reconstructs the original information (e.g., similarity matrix and edge matrix, etc.) from the latent space. 
When the reconstruction loss achieves a relative low level, we treat the output on the latent space as embedding vectors. 
\eat{In addition, metric learning is the other branch based on deep learning.  \cite{yang2019triplet,he2020sneq} developed a  Triplet Enhanced AutoEncoder (TEA)  which can make the embeddings equip  topological structure and preserve the discriminative information. All the mentioned methods before can only deal with single domain tasks  by deep graph convolutional network and have been studied for many years, but there are few works  started to handle multiple network, except ~\cite{inproceedings2020,zhang2019dane}. Specifically, DANE~\cite{zhang2019dane} first proposed to use domain-adaptive learned network to address multiple networks, where two shared weights GCNs were deployed to  encode source and target domain respectively and an extra discriminator network is to distinguish domain disparity to align two domain distributions. UDA~\cite{inproceedings2020}  proposed to preserve  local and global consistency fused by an inter-graph attention strategy into the embeddings and then leverage three classification loss  functions to categorize source domain, target domain and domain identification. AdaGCN~\cite{dai2019network} developed a semi-supervised framework, where two identical GCN is to  encode two domains into a common space and an adversarial domain adaptation component to reduce the distribution discrepancy between two domains.
}

\subsection{Learning to Hash}
Hashing as a powerful compression method that has been widely studied for many years due to its time and space efficiency~\cite{DBLP:journals/pami/GongLGP13,song2018binary}. We can generally divide hash methods into two categories: supervised and unsupervised. For the former, many works are dedicated to preserving pointwise similarity signals of raw data points into binary codes by various metrics, such as pairwise~\cite{ song2018binary,he2019one} and ranking list~\cite{wang2016deep}. As a matter of fact, for all variants of similarity calculation, the main purpose is to force the binary codes to be equipped with consistent semantic information. In comparison, unsupervised hash methods turn to pseudo similarity preservation~\cite{song2017deep} constructed by side information or the  model itself instead of 
directly using the ground truth. It is no doubt that supervised hash methods is significantly superior to the unsupervised ones in terms of the quality of hash codes.  
Another issue is that most of the existing methods~\cite{DBLP:journals/pami/GongLGP13,DBLP:conf/cvpr/Carreira-Perpinan15,song2017deep} can only perform well on single-domain data, that is, the learned hash function lacks the transferability between various datasets. To  alleviate this drawback,  cross-modal hash techniques~\cite{jiang2017deep,zhang2018attention} have been proposed to deal with multiple domains, such as images to text or text to images. 

\section{Problem Definition}
Let ${G^s} = \{(x_i^s,y_i^s)\}_{i = 1}^{{N^s}}$ denote the source domain, where {$y_i^s \in \{ 1,\ldots,N\}$} 
is the \emph{label} of $x_i^s$. ${G^t} = \{ x_i^t\} _{i = 1}^{{n_t}}$ denotes the unlabeled target domain. The goal of domain-adaptive hash learning is to  train a shared hash function $\mathcal{M}$ supervisedly on $G^s$ but unsupervisedly on  $G^t$ so that $\mathcal{M}$ can perform well on both domains.

\section{Methodology}

\begin{figure*}[]
	\centering
	\includegraphics[width=0.99\linewidth]{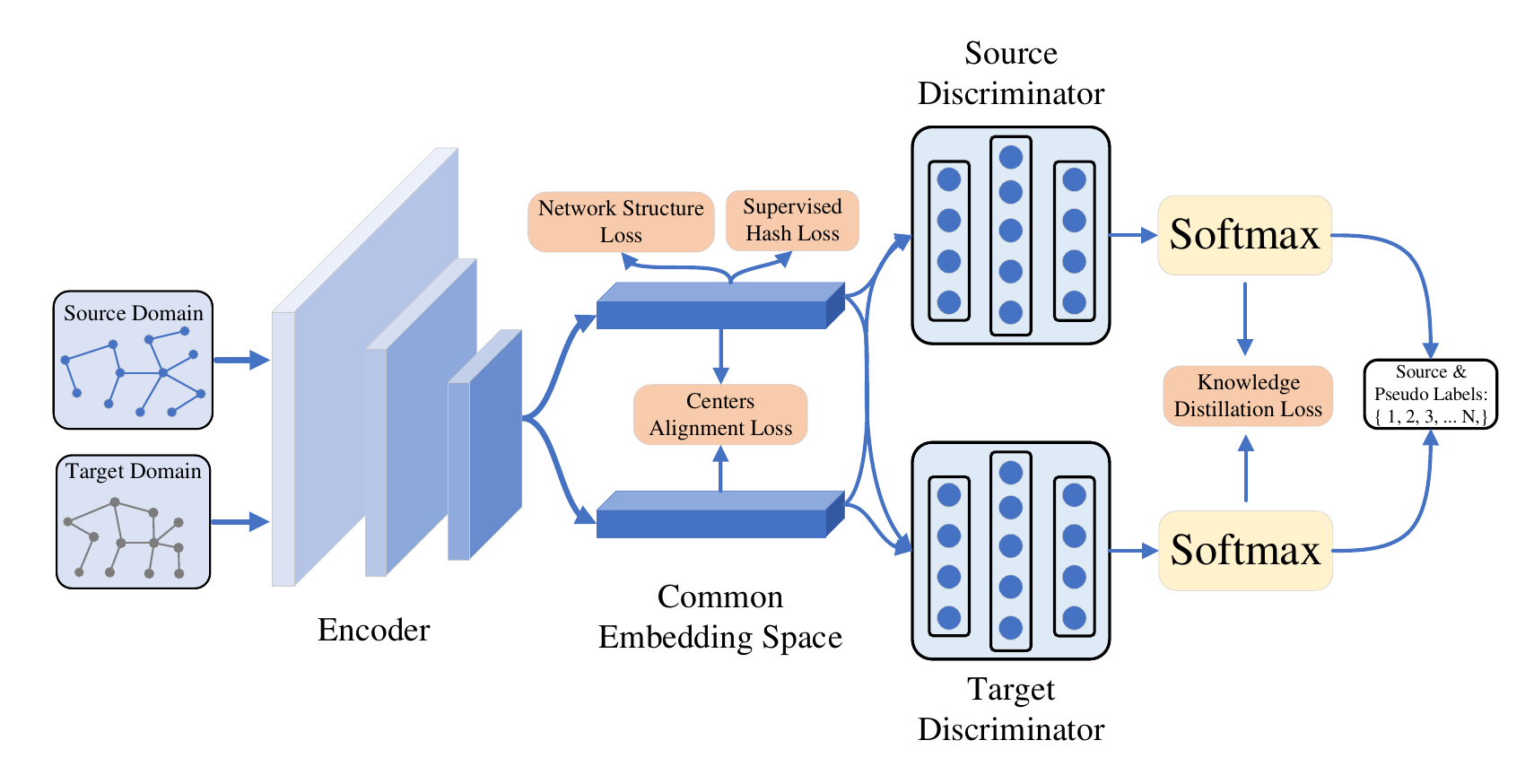}
	\caption{The overview of our framework, which consists of four components: (1)  deep encoder network, (2) differentiable supervised hashing, (3) semantic centers alignment, and (4) cross-domain intersected discriminators.}
	\label{fig.framework}
\end{figure*}

Figure~\ref{fig.framework} shows our overall framework, which consists of four modules: (1) the deep encoder network, (2) differentiable supervised hashing, (3) semantic centers alignment, and (4) cross-domain intersected discriminators. Specifically, the encoder aims at transforming input nodes' attributes into embeddings. The supervised hashing component focuses on learning a  hash function able to fit the target domain by supervision of the source domain. the cross-domain intersected discriminators are responsible for  transferring knowledge from source domain to target domain.   Finally, Semantic centers alignment constrains the semantics, e.g., cluster centers of the two domains, to be aligned.
\subsection{ Deep Encoder Networks}

Following a wide range of network embedding techniques~\cite{inproceedings2020,DBLP:journals/jmlr/GaninUAGLLML16}, we deploy the multi-layer perceptron (MLP)~\cite{DBLP:conf/iclr/BojchevskiG18} as our encoder network. Note that we use the same encoder network for both domains. 
First, we illustrate how our MLP-based encoder embeds network data into a latent space.
Concretely, our deep encoder network consists of multiple perceptron layers and takes nodes attributes $X$ as input. The encoding process for both domains can be formulated, as:
\begin{equation}
\begin{aligned}
h^{s}_i  &=  
\mathrm{MLP}(\bm{w}_i{h^{s}_{i-1}} + \bm{b}_i)\\
h^{t}_i  &=  
\mathrm{MLP}(\bm{w}_i{h^{t}_{i-1}} + \bm{b}_i)
\label{eq.encoder}
\end{aligned}
\end{equation}
where the $\mathrm{MLP}$ consists of three components: Dropout, LayerNorm  and  ReLU non-linearity; $i$ denotes the $i$-th layer of the multi-layer perceptron, $\bm{w}_i$ and $\bm{b}_i$ represent the $i$-th layer's weight and bias parameters respectively; and superscripts $^s$ and $^t$ denote the source and target domain respectively. It is worth noting that when $i=1$, $h^{s}_{0}$ (resp.\ $h^{t}_{0}$) is initialized by the source (resp.\ target) domain node features $x^s$ (resp.\ $x^t$). 

Mnih et al. \cite{mnih2013learning} has demonstrated that metric learning is effective in learning discriminative representations. Recently, many works~\cite{zhu2018deep,yang2019triplet} have proposed a triplet network based on metric learning to preserve neighborhood proximity into the latent space. 
Inspired by these work, we propose a hard contrastive loss to equip embeddings with nodes' neighborhood information. 
Specifically,  we define two types of node pairs: positive and negative, and consider two nodes as a positive pair $\mathcal{P}^+$ only if they have a direct connection, otherwise as a negative pair $\mathcal{P}^-$. Then we could formulate our objective function as:
\begin{equation}
\begin{aligned}
\mathcal{J}^s &= \frac{1}{N} \sum_{i=1}^{N}\max_{\bm{W}^e}(0,\lambda + \phi(z_i,p^+_j)-   \phi (z_i,p^-_j) ) 
\label{eq.dis1}
\end{aligned} 
\end{equation}
where $\lambda$ is a constant margin hyperparameter; $\bm{W}^e$ denotes the parameters of the encoder; and  $\phi(\cdot)$ is a function to measure the distance of two embeddings in the embedding space, for which we choose the Euclidean distance (L$_2$ norm). 
However, as the number of negative pairs are orders of magnitudes more than the number of positive pairs, the model is prone to inclining heavily to the negative samples, causing it to poorly preserve network structure.
 
To address this issue, we  propose {further}   to impose a hard contrastive loss on positive and negative \emph{groups} instead of pairs, i.e., our groupwise objective function aims at minimizing the maximal distance in the positive group whilst maximizing the minimal distance    in the negative group.
More concretely, the positive group of node $i$ is defined as all of its direct neighbors, denoted as $j \in \mathcal{P}_{i}^{+}$, while the negative group of node $i$ is those nodes not in the neighborhood of node $i$, that is, $j \in \mathcal{P}_{i}^{-}$.
Then, we could rewrite Eq.~(\ref{eq.dis1}) as the groupwise hard contrastive loss  as the below:
\begin{equation}
\begin{aligned}
\mathcal{L}_1 &= \frac{1}{N} \sum_{i=1}^{N}\max_{\bm{W}^e}(0,\lambda +\max_{j \in \mathcal{P}_{i}^{+}} \phi(z_i,p^+_j)-  \min_{j \in \mathcal{P}_{i}^-} \phi (z_i,p^-_j) )
\label{eq.dis}
\end{aligned} 
\end{equation}

In our implementation, since $|\mathcal{P}_i^+|$ is not too large, for a give anchor node $i$, we sample all its positive neighbors as  $\mathcal{P}_i^+$. On the other hand, due to the large size of $ |\mathcal{P}_i^-|$, we randomly sample  about {10$\times$$|\mathcal{P}_i^+|$} negative pairs to construct $\mathcal{P}_i^-$ each time. 

Compared  with Eq.~(\ref{eq.dis1}), Eq.~(\ref{eq.dis}) only adds the group constraint on the distance calculation, and it is natural to ask why it could perform better than Eq.~(\ref{eq.dis1}). We think the main reason is that Eq.~(\ref{eq.dis1}) is prone to falling into sub-optimality, because Eq.~(\ref{eq.dis1}) can only select $|\mathcal{P}_i^+|$ negative samples each time, but due to the large scale of  $|\mathcal{P}_i^-|$, it is hard to guarantee that all anchor nodes' positive pairs have smaller distances than its negative ones. By contrast, in Eq.~(\ref{eq.dis}), more negative pairs are sampled each time and we only select the negative points with the largest distance to optimize the contrastive loss, which benefits the model to find the optimal solution.

\subsection{Differentiable Supervised Hashing for Source Domain}
\label{sec:4.2}
In domain-adaptive learning, labels are available for the source domain, while it is often assumed that no label is available for the target domain~\cite{DBLP:conf/iccv/GopalanLC11,DBLP:journals/jmlr/GaninUAGLLML16}. Therefore, how to make use of the available labels plays a key role in domain-adaptive hashing learning. 
Intuitively, we adopt the supervised pairwise hash objective function in our preliminary work~\cite{he2019one} to preserve label similarity into hash codes, as below:
\begin{align}
\mathcal{L}_2 = \mathop {\min }\limits_{{\textbf{W}^e}} \frac{1}{2}{\sum\limits_{{s_{ij}} \in \mathcal{S}^s} {\left( \frac{1}{l}{b_i}{b_j}^\top - s_{ij}\right)} ^2}  
\label{eq.sgn}
\end{align}
where $\textbf{W}^e$ denotes parameters of the encoder, $b_i$, $b_j$ are hash codes generated by $b_i = \mathrm{sign}(z_i)$, $\mathcal{S}^s\in\{-1,1\}$ is a similarity matrix constructed from ground-truth labels of the source domain, and $l$ is the length of the hash code. Specifically, if two points have at least one same label, their similarity is defined as $1$ and otherwise $-1$. By optimizing Equation~(\ref{eq.sgn}), our model minimizes  the embedding distance of points with a same label but maximizes the distance across different labels. 

Unfortunately, as discussed before, the hash function $b_i = \mathrm{sign}(z_i)$ is not differentiable, so it cannot be minimized directly by backpropagation.  To solve this problem,  we treat the discretization as a classification problem, that is,  a hash code in each dimension is  the result of a binary classification, which is equivalent to adding {a binary classifiers for each discretization dimension}.  In the implementation, we add a linear classifier layer after the embeddings $z$, and the classified score is denoted as  $u_i \in \mathbb{R}^ {l\times k}$,  where $k$ is  set as  $2$ denoting the classified two options and   $l$ is the hash code length.
{For ease of illustration, we separate $z_i$ into $l$ blocks each of which is $k$-dimensional: i.e. $z_i=[z_{i1},z_{i2},\ldots,z_{i{l}}]$.} 
Inspired by Gumbel-Softmax~\cite{jang2016categorical,shu2017compressing,chen2020differentiable} that has gained great success in selecting categorical variables, we leverage it to estimate the gradient produced by our discretization, as below:
\begin{align}
&u_{i}=\left[ \mathrm{softmax}(\frac{z_{i1}+g_1}{\tau}), \mathrm{softmax}( \frac{z_{i2}+g_2}{\tau}),  \ldots,  \mathrm{softmax}( \frac{z_{i{l}}+g_{{l}}}{\tau}) \right]
\label{eq.5}\\
&\mathrm{softmax}( \frac{z_{ij}+g_j}{\tau})= \frac{\exp \left(\left(z_{ijd}+g_{d}\right) / \tau \right)}{\sum_{d=1}^{k} \exp \left(\left(z_{ijd}+g_{d}\right) / \tau \right)} 
\label{eq.6}
\end{align}
where $g_j \in \mathbb{R}^k$ is sampled from a Gumbel Distribution, $g_j = \mathrm{log}(-\mathrm{log}(U(0, 1))$ where $U$ is a uniform distribution and $\tau \in (0, \infty)$ is a temperature parameter to adjust the approximation~\cite{chen2018learning}.

Consequently, we can rewrite the discrete hash function (\ref{eq.sgn}) as:
\begin{align}
\centering
\mathcal{L}_2 = \mathop {\min }\limits_{{W^e}} \frac{1}{2}{\sum\limits_{{s_{ij}} \in S^s} {\left( \frac{1}{l}{u_i}{u_j}^\top - s_{ij}\right)} ^2}  
\label{eq.con}
\end{align}

Since Equation~(\ref{eq.con}) is differentiable, we can directly use gradient descent strategies to learn the approximate hash codes, where $u_i \approx b_i$. 
At the testing stage, we adopt  $\mathrm{argmax}(\cdot)$ to choose the maximum classified score's index as  hash codes, that is:
\begin{equation}
{b_i} = \left[\mathrm{argmax}(u_{i1}), \mathrm{argmax}(u_{i2}), \ldots, \mathrm{argmax}(u_{il})\right]
\end{equation}
 
\subsection{Cross-domain Intersected Discriminators} \label{sec:4.3}

The last section solves the discretization problem in hash learning. 
In this section, we describe how we preserve the semantics into binary hash codes to improve their quality.

As depicted in Figure~\ref{fig.framework}, our model consists of two discriminators, denoted as $D^s$ and $D^t$, aiming to classify  labels for the source domain and the target domain respectively.  Note that each discriminator consists of two $\mathrm{MLP}$ layers, followed by a classifier that produces a label prediction for each embedding.   Due to the fact that the source domain has rich-label information but the target domain is unlabeled, a key challenge is how to transfer the learned patterns under source-domain supervision to the target discriminator. To this end, many previous work~\cite{inproceedings2020,DBLP:conf/cvpr/TzengHSD17,Hu_2018_CVPR} used a pseudo-label strategy to generate proxy ground-truth annotations for the unlabeled domain in a self-supervised manner.  However, they simply  built two independent classifiers for each domain without too much mutual interaction except for letting the source classifier predict pseudo labels. We hypothesize that sharing information (knowledge) learned by both classifiers, instead of focusing on its own domain, will be more beneficial to the knowledge transfer between the two classifiers. 
Hence, we develop a cross-domain intersected discriminator component, that is, the source domain classifier classifies not only source domain data but also target domain data. 

First, the source discriminator classifies the source data by a cross-entropy loss under supervision:
\begin{equation}
\mathcal{L}_3^{{s}}=- \frac{1}{N^{s}} \min \limits_{\textbf{W}^e,\textbf{W}^s} \sum_{i=1}^{N^s} \mathbf{y}_{i}^s \log \left(\mathbf{\tilde{y}}_{i}^s\right)
\label{eq.9}
\end{equation}
where $\textbf{W}^s$ is the parameters of the source discriminator; $\mathbf{\tilde{y}}_i^s$ is the prediction probability from $D^s$ for the source domain point $i$; and $\mathbf{y}_i^s$ is the corresponding ground-truth label. 

For the unlabeled target domain, we first use the source discriminator to generate pseudo labels. 
To obtain highly precise pseudo labels, we set a threshold $T$ to select the target domain label, as shown below:
\begin{align}
{\mathbf{y}^{{t'}}_i} = \left\{ \begin{array}{l}
\arg\max ({\mathbf{\hat{y}}_i^t}){\rm{ }}~~~~~~~~~~~~~\text{if~} {\mathbf{\hat{y}}_i^t} > T\\
- 1{\rm{         }}~~~~~~~~~~~~~~~~~~~~~~~~\text{otherwise}
\end{array} \right.
\label{eq.threshold}
\end{align}
where $\mathbf{y}_i^{t'}$ denotes the pseudo label of the target domain point $i$, and $\mathbf{\hat{y}}_i^t$ is point $i$'s prediction probability from $D^s$. 
After that, we treat the pseudo labels as the proxy ground-truth and train the target-domain discriminator in a supervised way:
\begin{equation}
\mathcal{L}_3^{{t}}=-\frac{1}{N^{t}} \min \limits_{\textbf{W}^e,\textbf{W}^t} \mathbf{y}_i^{t'} \log \left(\mathbf{\tilde{y}}_{i}^t\right)
\label{eq.11}
\end{equation}
where $\textbf{W}^t$ is the parameters of the target discriminator, and $\mathbf{\tilde{y}}_i^t$ is the predicted label by $D^t$.

To enable the supervised signals to transfer to the target-domain discriminator, we further adopt the widely-used knowledge distillation strategy~\cite{tung2019similarity}. Specifically, we view the source domain as a teacher discriminator and the target discriminator as the student, aiming to mimic the prediction of the teacher and thus achieve the goal of reducing domain discrepancy. The knowledge distillation strategy for the student is formulated as follows:
\begin{equation}
 \mathcal{L}_{kl} =  \frac{1}{N^{s}}\mathop {\min }\limits_{{\textbf{W}^e},\textbf{W}^s,\textbf{W}^t}\sum_{i} \mathrm{KL}\left(\hat{\mathbf{y}}_{i}^{t} \| \tilde{\mathbf{y}}_{i}^{t}\right)
 \label{eq:kl}
\end{equation}
where  $\mathrm{KL}(\cdot)$ denotes the Kullback-Leibler divergence to measure the discrepancy between the two domains. 

\subsection{Semantic Centers Alignment}
Distributional alignment plays a vitally important role in the classification of the target domain. If both domains are well aligned, the knowledge learned on the source domain, especially the discriminative information learned  by the supervised hash component, can readily transfer to the  target domain and boost its classification performance. Although many approaches~\cite{zhang2019dane,xie2018learning} have proposed to use a generative adversarial network (GAN) to handle this problem, they do not make sure that the semantic subspaces in the latent space is regionally aligned. 

Therefore, we propose a semantic centers alignment component to force the two domains to semantically align in terms of each label's cluster center, that is, the same class centers of the two domains should be close to each other. It is true that if centers of both domains are highly aligned, the discriminative clues learned by supervised signals of the source domain can be readily transferred to the target-domain embeddings.
The critical challenge is how to measure the alignment of the two domains, since the target domain is unlabeled. To address this problem, we exploit a semantic centers alignment loss based on clustering algorithms, i.e.\ the K-means algorithm, to calculate each category's center and force the same class centroid to be close. The semantic centers alignment loss can be formulated as:
\begin{equation}
{\mathcal{L}_4} = \mathop {\min _{\bm{W}^e}} (\sum\limits_{i = 1}^K {\varphi (\frac{1}{{{m_i}}}\sum\nolimits_{{y^s} = i} {{z^s}} ,\frac{1}{{{n_i}}}\sum\nolimits_{{y^{ t'}} = i} {{z^t}} )} )
\label{eq:ls}
\end{equation}
where $K$ is the number of classes and $m_i$ (resp.\ $n_i$) denotes the number of samples in the same cluster in the source (resp. target) domain. $\varphi(\cdot,\cdot)$ is the function that measures the distance of different centers. 
$y^{t'}$ is the pseudo label generated by Equation~(\ref{eq.threshold}). 
In this work, we leverage the Euclidean distance to define the distance between centers, i.e., $\varphi ({z_i},{z_j}) = {\left\| {{z_i} - {z_j}} \right\|^2}$. 
Additionally, due to the high time cost when calculating K-means on all samples,  we do not precisely calculate the centers of all nodes in the training stage, but instead calculate the centers of mini-batches to approximate the global centers. It is worth noting that we usually set a large batch size to make sure the K-means clustering covers all categories  during training.

\section{Theoretical Analysis}

In the previous section we have described a suite of techniques to reduce two domains' distribution disparity, including cross-domain discriminators and semantic centers alignment. It is important to discern why they are effective to reduce that discrepancy for domain-adaptive hashing. To answer this question, we theoretically analyze the latent correlation between their distribution alignment and objective loss functions. In this work, there are two types information to be embedded: neighborhood structure proximity and node label semantics. In the remainder of this section we take the latter as an example to elucidate the effectiveness.

Let our learned domain-adaptive hash function be denoted as $\mathcal{F}$ and all required parameters as $\vartheta$, we could formulate a hash code of a source node $x^s$ as: $\mathcal{F}(x^s;\vartheta)$. Similarly, given a target domain node $x^t$, its generated hash code is denoted as $\mathcal{F}(x^t;\vartheta)$. The loss function for the source domain can be expressed as:

\begin{equation}
	L_{src} =  \sum_{x_i^s \in G^s} \mathcal{H}(v_i^s,\mathcal{F}(x_i^s;\vartheta))
	\label{eq:l_src}
\end{equation}
where $v_i^s$ denotes the group-truth hash vector of $x^s$ and  $\mathcal{H}(\cdot,\cdot)$ is to calculate a hamming distance between two hash codes. {It is worth noting that $v_i^s$ does not exist in our dataset, but can be obtained by our supervised hash component~\ref{sec:4.2}.} 

Similarly, we could express the loss function on the target domain $G^t$ as below:
\begin{equation}
L_{tgt} =  \sum_{x_i^t \in G^t} \mathcal{H}(v_i^t,\mathcal{F}(x_i^t;\vartheta))
\label{eq:l_tgt}
\end{equation}
where $v^t_i$ denotes the  learned hash codes on target domain. 

Then we will have the following hypothesis:
\begin{equation}
   L_{tgt} - L_{src} \leq \sum_{x_i^t \in G^t ~\& ~x_i^s \in G^s } \mathcal{H}( \mathcal{F}(x_i^t;\vartheta),\mathcal{F}(x_i^s;\vartheta)  )
\end{equation}
\\ 

\begin{proof}
\begin{equation}
  L_{tgt} - L_{src} =  \sum_{x_i^t \in G^t} \mathcal{H}(v_i^t,\mathcal{F}(x_i^t;\vartheta)) - \sum_{x_i^s \in G^s} \mathcal{H}(v_i^s,\mathcal{F}(x_i^s;\vartheta)) 
  \label{eq:lt-ls}
\end{equation}

If $x^t_i$ and $x^s_i$ have a same node label, we view both of them as having the same  hash codes, i.e., $v_i^t=v_i^s$. Note that $v^t_i$ can be acquired by our cross-domain discriminators~\ref{sec:4.3}. Hence, we could reorder Eq.~(\ref{eq:lt-ls}) so that $v_i^t=v_i^s$ for each pair ($x^t_i$, $x^s_i$). Note that here we consider $|G^t|=|G^s|$ and each class has the same number of points. If $|G^t|$ is not equal to $ |G^s|$, we could solve it by a resampling strategy~\cite{good2006resampling} to force $|G^t| = |G^s|$, and the same treatment for each category. 

Then, we could rewrite Eq.~(\ref{eq:lt-ls}) in an alignment way as:
\begin{equation}
L_{tgt} - L_{src} =  \sum_{x_i^t \in G^t} \mathcal{H}(v_i,\mathcal{F}(x_i^t;\vartheta)) - \sum_{x_i^s \in G^s} \mathcal{H}(v_i,\mathcal{F}(x_i^s;\vartheta)) 
\label{eq:lt-ls2}
\end{equation}
where $v_i$ denotes their aligned   hash codes. 

 Since the Hamming distance $\mathcal{H}$ satisfies the triangular inequality:
  \begin{equation}
  	\mathcal{H}(A,B) \leq \mathcal{H}(A,C)+\mathcal{H}(C,B)
  \end{equation}
 Hence, we could obtain the following inequality: 
 \begin{equation}
 \begin{aligned} 
 	 L_{tgt} - L_{src} &= \sum_{x_i^t \in G^t} \mathcal{H}(v_i,\mathcal{F}(x_i^t;\vartheta)) - \sum_{x_i^s \in G^s} \mathcal{H}(v_i,\mathcal{F}(x_i^s;\vartheta)) \\
 	 & \leq  \sum_{x_i^t \in G^t ~\& ~x_i^s \in G^s } \mathcal{H}( \mathcal{F}(x_i^t;\vartheta),\mathcal{F}(x_i^s;\vartheta)  )
 \end{aligned}
 \label{eq:upper}
 \end{equation}

 From Eq.\ref{eq:upper}, we could know that when we optimize $L_{src}$ and  the Hamming distance of the two domain's hash codes, i.e., reducing the disparity of the two domains, $L_{tgt}$ is also being optimized. 

\end{proof}
 
\section{Learning}

In summary, our model consists of the following five main objective functions: network structure embedding loss $\mathcal{L}_1^*$, semantic hashing loss $\mathcal{L}_2$ (Equation~(\ref{eq.con})), cross-domain classification loss $\mathcal{L}_3^*$, KL loss $ \mathcal{L}_{kl}$ (Equation~(\ref{eq:kl})) and semantic centers alignment loss $\mathcal{L}_4$ (Equation~(\ref{eq:ls})). 
The overall objective function is:
\begin{align}
\mathcal{L} = \mathop {\min _ {\bm{W}^e, \bm{W}^s, \bm{W}^t} } (\alpha \mathcal{L}_1^* + \beta \mathcal{L}_2 + 
\sigma \mathcal{L}_3^* + \delta \mathcal{L}_4^a+  \mathcal{L}_{kl})
\label{all_eq}
\end{align}
where the superscript $^*$ denotes the loss term for the source domain ($s$) and target domain ($t$) respectively for simplicity, and $\alpha$, $\beta$, $\sigma$ and $\delta$ are four hyper-parameters to balance the structure, hashing,  cross-domain classification, and alignment loss terms, respectively. Since all parameters $\bm{W}^e, \bm{W}^s, \bm{W}^t$ are continuous, we can directly use stochastic gradient descent (SGD) to optimize all the parameters. 

For the centers alignment component, since both domains' center calculation is based on mini-batches, it makes sense that the larger the mini-batch size is, the more accurate cluster centroids can be obtained. Moreover, during the training stage, we need to ensure the batch size is much larger than the number of classes ($K$).  At the same time, the semantic centers $\mathcal{\bm{C}}$ are also learned by the following update strategy:
\begin{align}
\mathcal{\bm{C}}_r^* &= \epsilon \mathcal{\bm{C}}_{r-1}^* +(1- \epsilon)   \Phi(\mathbf{z}^*)
\label{eq:snl}
\end{align}
where the superscript $^*$ represents the domains $s$ or $t$, $r$ denotes the $r$-th mini batch, $\mathcal{\bm{C}}$ denotes the centers calculated by Equation~(\ref{eq:snl}), $\epsilon$ (set as $0.3$ in our experiments) is the update step size, and $\Phi(\cdot,\cdot)$ is the clustering function K-means to calculate each class' center in the $r$-th mini batch.

%

\section{Experiments}

In this section we will evaluate our domain-adaptive hashing model against some state-of-the-art models on networks. We begin by describing the benchmark datasets, baseline models and implementation details. 

%
%
%
%
%

\subsection{Datasets}

%
%
%

Following UDA~\cite{inproceedings2020}, we  conduct our experiments on three citation networks obtained from ArnetMiner~\cite{tang2008arnetminer}. Brief statistics of the three networks are shown in Table~\ref{tab.datasets1}. We sample subsets from three large citation networks: DBLPv4 (D), ACMv8 (A) and Citationv1 (C). To reduce the overlap between different datasets, we extract published papers from different periods for these three datasets following UDA~\cite{inproceedings2020}. 
Papers are classified into eight categories: Engineering, Electronic, Software Engineering, Mathematics, Theory, Applied, Artificial Intelligence , and Computer Science. For the attributes, we extract the word frequency of each paper's abstract, which is represented as an $8,328$-dimensional vector.  


\begin{table}[]
	\caption{Brief statistics of the three datasets.}
	\centering
	\label{tab.datasets1}
	\begin{tabular}{l*{4}{r}}
		\toprule
		$G$  & 
		DBLPv4   & ACMv8 &     Citationv1      \\ \midrule
		$|V|$    & 6,209 & 8,173 & 4,350 \\ 
		$|E|$    & 8,056 & 22,753 & 8,513 \\ 
		Attr.    & 8,328 & 8,328 & 8,328\\ 
		Labels   & 8 & 8  & 8  \\ 
		\bottomrule
	\end{tabular}
\end{table}

\subsection{Baselines}

We choose the following state-of-the-arts discrete hash methods for network embeddings as our baselines. 
\begin{itemize}
	\item  SH~\cite{datar2004locality} is a classical and   widely-applied learning to hash method for the approximate nearest neighbour search task.
	
	\item Discrete Collaborative Filtering (DCF)~\cite{zhang2016discrete}  is a principled hashing method able to tackle the challenging discrete optimization problem in hash learning and avoid large quantisation errors caused by two-step optimization.
	
	\item DNE~\cite{shen2018discrete} is the first work to a discrete
	representation for networks by preserving Hamming similarity.
	
	\item NetHash~\cite{wu2018efficient} utilises the randomized hashing technique to embed trees in a graph, which can preserve information closer to the root node as much as possible.
	
	\item Binarized Attributed Network Embedding (BANE)~\cite{yang2018binarized} develops  a Weisfeiler-Lehman proximity matrix
	that can preserve the dependence between node attributes and connections via combining the features from neighbouring nodes. 
	
	\item Information Network Hashing (INH) \cite{lian2018high} is an embedding compression method based on matrix factorization and able to  preserve high-order proximity into binary codes.
	\item Discrete Embedding for Latent Networks (DELN) \cite{yang2020discrete} is an end-to-end discrete network embedding method to  learn binary representations.

\end{itemize} 

\subsection{Implementation Details}

 The feature encoder network consists of three $\mathrm{MLP}$ modules  and the last layer outputs $256$-dimensional embeddings. The batch size is set to $400$, which is much greater than the number of labels, so it is safe to guarantee effectiveness of the K-means algorithm. For a fair comparison with baseline methods, all methods' hash code length is set to $128$. 
 The temperature $\tau$ in Equation (\ref{eq.6}) is set to 1.
The learning rate, and the five hyperparameters $\alpha$, $\beta$, $\delta$, $\delta$ and $\lambda$ are set to $0.005$, $1.0$, $0.01$, $1.0$, $0.1$ and $5$ respectively, and are obtained by grid search on the validation set. The threshold $T$ for pseudo label selection is set to $0.85$. 

For the evaluation on the node classification task, we first generate all nodes' embeddings and then train a one-vs-rest logistic regression classifier to classify the embeddings, where all methods use the same-dimensional hash codes for training and testing. We measure the mean score of Micro F1 and Macro F1 metrics to evaluate the performance of node classification, following DANE~\cite{zhang2019dane}, and use the area under curve (AUC) score to evaluate the performance of link predication, following Graph2Gauss~\cite{DBLP:conf/iclr/BojchevskiG18}. For link prediction, we randomly select $5$\% and $10$\% edges as the validation and test set respectively, following Graph2Gauss~\cite{DBLP:conf/iclr/BojchevskiG18}. 


All experiments were performed on a workstation with $256$ GB memory, $32$ Intel(R) Xeon(R) CPUs (E$5$-$2620$ v$4$ @ $2.10$GHz) and $8$ GeForce GTX $1080$Ti GPUs.  


\subsection{Cross-domain Node Classification Results }
Node classification is a standard task to evaluate the embedding performance of preserving semantics. In this work, we also adopt it to test the discrete hash codes' capability to learn semantics. 

Table~\ref{tab.hash} shows the node classification results (of discrete embeddings) compared with the state-of-the-art discrete network embedding methods. For a fair comparison, due to the fact that SH cannot learn representations for each node, we use the feature learned from UDAH to train SH, i.e., the latent embeddings $z$, while the other models use nodes' attributes to train. 

From Table~\ref{tab.hash}, it can be observed that our method UDAH achieves the best performance over all but one domain-transfer tasks, except for C$\rightarrow${A} with DELN being 0.33 percentage points higher than ours. On average, UDAH is superior to the baseline methods, surpassing the second best method DELN by 2.42 percentage points. 

We consider that the main reason is the compared methods can only work well on single domains, but perform poorly on new domains with relatively large distribution disparity.  For example, the majority of discrete embedding methods, including DNE, INH and DELN, leverage a matrix factorization technique to learn the binary codes. They decompose the input attribute matrix into hash codes with the constraint to reconstruct the original attribute matrix. Their common issue lies in the fact that the target network has a large different attributes distribution from the source network, which results in large reconstruction errors by the matrix decomposition operation, and finally leading to poor results on the domain transfer. In contrast, UDAH explores several techniques to mitigate the problem of attribute distribution shift, including the KL divergence loss and semantic centers alignment. 
\begin{table}[]
	\centering
	\caption{
		Node classification results on six cross-domain tasks compared with the state-of-the-art discrete embedding methods in terms of the mean of Micro-F1 and Macro-F1 score (\%).}
	\label{tab.hash}
	\begin{tabular}{lccccccr}
		\toprule
		Methods & A$\rightarrow$D & A$\rightarrow$C & C$\rightarrow$D & C$\rightarrow$A & D$\rightarrow$A & D$\rightarrow$C & Average \\\midrule
		SH     & 21.25 & 16.51 & 20.37 & 17.52 & 19.68 & 19.34 & {19.13} \\
		DCF    & 20.47 & 18.13 & 21.52 & 19.11 & 20.24 & 23.06 & {20.42} \\
		NetHash& 23.83 & 19.38 & 24.15 & 23.05 & 23.52 & 24.28 & {23.06} \\
		DNE    & 23.51  & 24.81 & 22.38 & 21.91 & 22.40 & 25.06 & {23.35} \\
		
		BANE   &  25.72 & 22.15  & 20.63  & 22.40  & 23.08  &  20.75 & {22.46} \\
		INH & 21.40 & 26.14 & 24.36 & 25.60 & 24.41 & 25.40 & {24.55} \\
		
		DELN & 25.06 & 26.83 & 28.59 & \textbf{27.12} & 26.13 & 25.23 & {26.48} \\
		
		\textbf{UDAH}  &   \textbf{25.41} & \textbf{29.82}& \textbf{31.37} & 26.79 & \textbf{29.25} & \textbf{30.14} & \textbf{28.90}\\ \bottomrule
	\end{tabular}
	
\end{table}
 In the ablation study described later, we will further test the performance of each component to analyze how much they contribute to domain-adaptive hash learning. It is worth noting that SH gains  comparable results with DCF and does not show catastrophic performance degradation as we originally anticipated, possibly because SH is trained by our learned continuous representations, which, to some extent, confirms that our network embedding strategy can learn high-quality embeddings for multiple domains.
 
 Therefore, we can conclude that our UDAH handle domain-adaptive hash learning more effectively than the conventional single-domain discrete network embedding methods in terms of semantics preservation.

\subsection{Link Prediction Results}

Link prediction evaluates the learned hash codes' ability to reconstruct the original network's neighbor structure. 
Following  Graph2Gauss~\cite{DBLP:conf/iclr/BojchevskiG18}, the validation/test set consists of 5\%/10\% of edges randomly sampled from the network respectively, and we randomly selected edges and an equal number of non-edges from the test set. Table~\ref{tab.hash_link} shows the link prediction results of discrete embeddings on the six cross-domain tasks. 
 \begin{table}[]
 	\centering
 	\caption{
 		Link prediction results on six cross-domain tasks compared with the state-of-the-art discrete embedding methods in terms of the mean of AUC score (\%).
 	}
 	\label{tab.hash_link}
 	\begin{tabular}{lccccccr}
 		\toprule
 		Methods & A$\rightarrow$D & A$\rightarrow$C & C$\rightarrow$D & C$\rightarrow$A & D$\rightarrow$A & D$\rightarrow$C & Avg. \\ \midrule
 		
 		SH & 65.24 & 67.51 & 64.37 & 61.82 & 62.43 & 65.29 & 64.44 \\
 		DCF & 68.47 & 66.38 & 65.12 & 63.22 & 67.59 & 64.18 & 65.83 \\
 		NetHash & 71.64 & 69.73 & 68.51 & 64.14 & 68.65 & 67.74 & 68.40 \\
 		DNE & 65.81 & 70.34 & 69.15 & 70.35 & 71.19 & 68.31 &  69.19 \\
 		BANE & 66.37 & 71.62 &  65.79 & 69.42 & 66.36 & 66.67 & 67.71  \\
 		INH & 74.51 & 76.14 & 70.38 & 69.60 & 72.43 & 72.40 & 72.31 \\
 		DELN &  70.29 & 68.41  & 65.82 &  67.31 & 66.81 & 68.03 &  67.78 \\
 		\textbf{UDAH} &  \textbf{77.02} &  \textbf{78.52} &  \textbf{74.27} &  \textbf{71.79} & \textbf{75.37} & \textbf{74.36} & 
 		\textbf{75.17}
 		\\ \bottomrule
 	\end{tabular}
 \end{table}
 
From the results we can observe that our UDAH method exceeds other single-domain discrete embedding methods over all domain transfer tasks. In particular, UDAH is $2.86$ points better than the second best method INH. Since INH places emphasis to preserving high-order proximity, it shows superiority to other matrix factorization based models such as DNE. 
It is worth noting that DELN achieves poor performance in this task, although it obtains competitive results in node classification. 
  
In summary, although single-domain embedding methods can handle unitary networks, they consistently show performance decreases when evaluated in a domain-adaptive setting, in which our method achieves a substantial performance advantage due to techniques specifically designed to tackle this problem. 

\subsection{Node Recommendation}
Node recommendation is a widely employed task to evaluate retrieval performance for social and commercial networks, for which discrete embeddings can save much time. Given a query hash code, node recommendation aims to returning a list of nodes, ranked by their structural similarity. Following the settings in INH~\cite{lian2018high}, we sample 90\% of neighbours of each node  to  train the model while the remaining 10\% neighbours are reserved for evaluation, and use NDCG@50 as the evaluation metric. 

Table \ref{tab.recom} presents the performance of node recommendation on six cross-domain tasks. From the table, we could observe that our UDAH method outperforms all the baseline methods in terms of the average performance, outperforming the second best method INH by approx.\ 2 points. INH in turn outperforms he third best DELN by 1.26 points. Although both of them are based on matrix factorization, INH's advantage comes from its capability to learn high-order proximity. Although UDAH only explores the first-order neighborhood structure preservation, our other techniques aiming to reduce the domain discrepancy play an important role in network structure and semantics preservation. The other methods, such as DCF, NetHash and DNRE, perform much more poorly in this task.
\begin{table}[]
	\centering
	\caption{
		Node recommendation results on six cross-domain tasks compared with the state-of-the-art discrete embedding methods in terms of NDCG@50 (\%).
	}
	\label{tab.recom}
	\begin{tabular}{lccccccr}
		\toprule
		Methods & A$\rightarrow$D & A$\rightarrow$C & C$\rightarrow$D & C$\rightarrow$A & D$\rightarrow$A & D$\rightarrow$C & Avg. \\ \midrule
		
		SH         &  9.64 &  12.74 & 14.04  & 13.16 &  10.72 & 9.12  &  11.57 \\
		DCF        &  11.61 & 14.93 & 17.38  & 15.47 &  13.58 & 10.36 &  13.89\\
		NetHash    &  14.29 & 13.07 & 16.62  & 14.70 &  12.93 & 12.03 &  13.94 \\
		DNE        &  13.84 & 15.39 & 18.85  & 14.52 &  14.02 & 14.28 &  15.15  \\
		BANE       &  19.30 & 16.83 & 19.40  & 18.31 &  15.19 & 17.95  &  17.83  \\
		INH        &  20.51 & 22.34 & 26.31  & 21.38 &  18.30 & \textbf{20.76} & 21.60  \\
		DELN       &  19.73 & 20.15 & 24.25  & 20.17 &  17.84 & 19.31 &    20.24\\
		\textbf{UDAH} & \textbf{23.12} & \textbf{24.53}  & \textbf{28.10}  & \textbf{24.28}  & \textbf{21.15} & {20.23} & \textbf{23.56} \\\bottomrule
	\end{tabular}
\end{table}

\subsection{Ablation Study}

In this section, we aims to study the effectiveness of each component in our model. 
Specifically, for the five components: groupwise contrastive loss, Guambel-softmax strategy, cross-domain classification loss, KL divergence loss, and semantic center alignment loss, we ablate them into the following five model variants: 
 \begin{itemize}
 \item $\mathcal{-}\mathcal{L}_1$ uses a point-wise contrastive loss Equation (\ref{eq.dis1}) instead of  our group-wise  constraint Equatin (\ref{eq.dis}).

\item $\mathcal{-}\mathcal{L}_2$ uses the conventional hash learning strategy in \cite{he2019one} instead of the Gumbel-Sofmax strategy in Equations (\ref{eq.5}) and (\ref{eq.6}).

\item $\mathcal{-}\mathcal{L}_3$ removes the cross domain classification loss, i.e.\ Equation (\ref{eq.9}) and (\ref{eq.11}).

\item $\mathcal{-}\mathcal{L}_{kl}$ discards the KL divergence loss, i.e.\ Equation (\ref{eq:kl}).

\item $\mathcal{-}\mathcal{L}_{4}$ removes the semantic centers alignment loss, i.e.\ Equation (\ref{eq:ls})
\end{itemize}

In addition, we further modify UDAH into a none-domain-adaptive version, dubbed $\mathrm{NoDAH}$, which has {  $ \mathcal{L}_1$,  {$ \mathcal{L}_2$} 
 and $\mathcal{L}_3^s$ } components.

We thoroughly conduct a wide range of experiments on node classification and link prediction, as shown in Tables~\ref{tab.cls} and~\ref{tab.lp}, respectively. It is worth noting that all variants are under the same experimental configurations expect for their corresponding ablated module(s). From the two tables, we could make the following observations:
 
 (1) $\mathrm{NoDAH}$ performs the worst among all the variants, due to the remove of all domain-adaptive learning strategies, such as  Kullback–Leibler divergence loss, cross-domain discriminator, and semantic centers alignment. Hence, it is reasonable that $\mathrm{NoDAH}$ achieves comparable results with other single-domain discrete embedding methods such as DCF. 
 
 (2) From the comparison between $\mathcal{-}\mathcal{L}_1$ and UDAN, a slight decrease can be seen in the variant of $\mathcal{-}\mathcal{L}_1$, which confirms that our groupwise contrastive loss is more effective than the pairwise version. Besides, we could observe that the groupwise contrastive loss plays a more important role in link prediction than in node classification, possibly because $\mathcal{L}_1$ aims to preserve network structure instead of semantics.
 
 (3) Comparing  $\mathcal{-}\mathcal{L}_2$ with UDAH, we could notice that the Gumbel-Softmax reparameterization trick brings about 2 and 1 points lift on node classification and link prediction, respectively. We conjecture the reason is that Gumbel-Softmax completely discards the discretization function $\mathrm{sign}(\cdot)$ during training and test by multiple linear classifiers to produce hash codes, which bypasses relaxation errors generated by $\mathrm{sign}(\cdot)$ or $\mathrm{tanh}(\cdot)$ functions. 
 
(4) Our three proposed techniques that aim at transferring the knowledge supervisedly leaned on the source domain to the target domain, i.e.\ cross-domain discriminators Eq.~(\ref{eq.9}) and (\ref{eq.11}), knowledge distillation loss Eq.~(\ref{eq:kl}), and semantic centers alignment Eq.~(\ref{eq:ls}), all contribute positively to the final performance. Specifically, among the three strategies, the cross-domain discriminators and center alignment are the more important and have larger impact on performance, while knowledge distillation seems to be not as critical as them. 

In summary, all proposed components can make positive task-specific contributions to domain-adaptive hash learning. Moreover, the cross-domain discriminators and center alignment modules play the most important role in UDAH. 

In the next section, we will qualitatively evaluate the embeddings produced by UDAH by visualizing the learned embeddings under different settings.

\begin{table}[]
	\centering
	\caption{The impact of  each component on the task of node classification (\%).}
	\label{tab.cls}
	\begin{tabular}{l*{8}{c}}
		\toprule
		Methods & A$\rightarrow$D & A$\rightarrow$C & C$\rightarrow$D & C$\rightarrow$A & D$\rightarrow$A & D$\rightarrow$C & Avg. \\
		\midrule
		
		NoDAH & 20.62 & 21.30 & 23.17 & 19.45 & 21.35 & 23.43 & 21.55  \\ 
		$\mathcal{-}\mathcal{L}_1$ & 23.91 &  26.49 & 29.03 & 25.17 & 28.11 & 27.84 &  26.76 \\  
		$\mathcal{-}\mathcal{L}_2$ & 22.53 &  25.20 & 27.42 & 26.20 & 27.26 & 28.01 & 26.10 \\ 
		
		$\mathcal{-}\mathcal{L}_3$ & 23.18 &  24.00 & 26.07 & 24.76 & 24.20 & 25.25 & 24.58   \\ 
		$\mathcal{-}\mathcal{L}_4$ & 22.42 &  23.51 & 25.71 & 23.35 & 23.85 & 26.72 &  24.19   \\ 
		$\mathcal{-}\mathcal{L}_{kl}$ & 24.28 & 26.98 & 28.87 & 24.69 & 27.59 & 28.12 & 27.76   \\ 
	
		\textbf{UDAH} &   {25.41} & {29.82}& {31.37} & 26.79 & {29.25} & {30.14} & {28.90} \\ 
		\bottomrule
	\end{tabular}
\end{table}

\begin{table}[]
\centering
	\caption{The impact of  each component on the task of link prediction (\%).}
	\label{tab.lp}
	\begin{tabular}{l*{8}{c}}
		\toprule
		Methods & A$\rightarrow$D & A$\rightarrow$C & C$\rightarrow$D & C$\rightarrow$A & D$\rightarrow$A & D$\rightarrow$C & Avg. \\
		\midrule
		$\mathrm{NoDAH}$  & 66.14 & 65.34 & 66.80 & 64.19 & 65.72 & 64.39 & 65.44 \\  
		$\mathcal{-}\mathcal{L}_1$ & 74.34 & 76.11 & 70.51 & 67.25 & 71.30 & 72.04 & 71.95 \\  
		$\mathcal{-}\mathcal{L}_2$ & 76.82 & 78.03 & 73.25 & 70.10 & 74.24 & 74.18 & 74.43 \\ 
		$\mathcal{-}\mathcal{L}_3$ & 75.95 & 76.24 & 72.06 & 68.17 & 72.60 & 72.45 & 72.92  \\ 
		$\mathcal{-}\mathcal{L}_4$ & 76.42 & 75.91 & 71.82 & 69.02 & 71.25 & 70.29 & 72.45  \\ 
 	$\mathcal{-}\mathcal{L}_{kl}$  & 75.11 & 77.50 & 73.03 & 71.20 & 74.62 & 73.25 & 74.12   \\ 
		\textbf{UDAH}   &  {77.02} &  {78.52} &  {74.27} &  {71.79} & {75.37} & {74.36} & 
		{75.17}\\ 
		\bottomrule
	\end{tabular}
\end{table}

 \begin{figure*}[]
 	\centering
 	
 	\subfloat[]
 	{        \includegraphics[width=0.3\linewidth ]{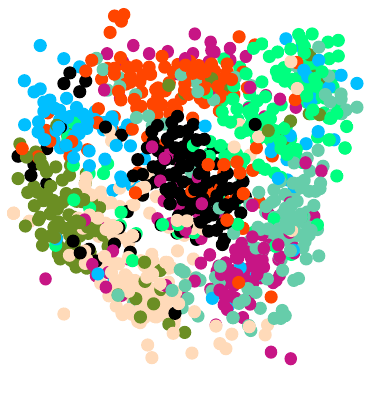}\label{fig.a}
 	}
 	\hspace {0.1cm}
 		\subfloat[]
 	{        \includegraphics[width=0.3\linewidth ]{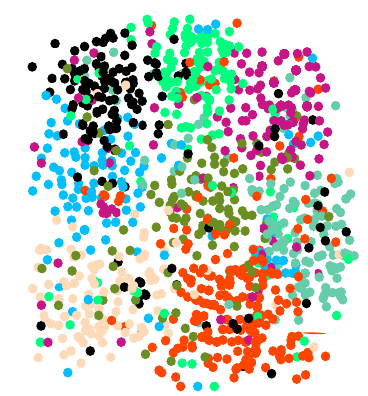}\label{fig.b}
 	}
 	\hspace {0.1cm}
 	\subfloat[ ]
 	{
 		\includegraphics[width=0.3\linewidth ]{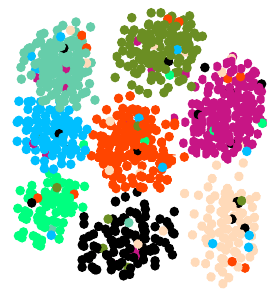}
 		\label{fig.c}
 	}
 	\caption{A visualization of the target domain embeddings learned with and without domain adaptation.  ($a$) is only trained by  NoDAH and ($b$) is trained by UDAH, from DBLPv4$\to$Acmv8.}
 	\label{fig.feat_vis}
 \end{figure*}

 \subsection{Embedding Space Visualization}
 
Feature visualization is a common way to evaluate the quality of the learned embeddings. Thus, we use t-SNE to reduce the embeddings to $2$-dimensional vectors and randomly sample some embeddings. Figure~\ref{fig.feat_vis} shows the visualization results of $\mathrm{NoDAH}$ and UDAH on a graph with $1,000$ nodes randomly selected from ACMv8, where the three models are trained on DBLPv4 and colors represent node classes. 
 
Figure~\ref{fig.a}, \ref{fig.b} and \ref{fig.c} visualizes the embeddings of NoDAH,  DELN and  UDAH, respectively. Generally, we can make the observation that  nodes of the same labels tend to cluster together in Figure~\ref{fig.c} where  nodes of different labels are separated by a large margins. However, the points in Figure \ref{fig.a} and \ref{fig.b} are much more tightly tangled, which confirms that single-domain network embeddings do not perform well in domain-adaptive settings. On the other hand, our model UDAH can 
 effectively learn discriminative embeddings in domain-adaptive transfer learning.


\section{Conclusion}
In this paper, we address the domain-adaptive learning to hash problem for networks, which is an under-explored but important problem that facilitates network analysis tasks in a time- and space-efficient manner. 
We develop UDAH, an end-to-end unsupervised domain-adaptive hash method to learn adaptive discrete embeddings. Specifically, we propose a suite of techniques: groupwise contrastive loss to preserve network structure, cross-domain discriminators and knowledge distillation modules to transfer knowledge of the source domain to the target domain, and center alignment to reduce the distribution disparity of both domains. 
Evaluation on three benchmark datasets against a number of state-of-the-art learning to hash methods demonstrate the superiority of UDAH on three tasks: node classification, link prediction and node recommendation. 
In future, we plan to extend our framework to support more diverse adaption scenarios, such as the transfer from social networks to citation networks.

 \bibliographystyle{elsarticle-num} 
 \bibliography{cas-refs}

\end{document}